\documentclass[conference]{IEEEtran}
\IEEEoverridecommandlockouts

\usepackage{cite}
\usepackage{amsmath,amssymb,amsfonts}
\usepackage{algorithmic}
\usepackage{graphicx}
\usepackage{textcomp}
\usepackage{xcolor}
\usepackage{booktabs}
\usepackage{float}

\def\BibTeX{{\rm B\kern-.05em{\sc i\kern-.025em b}\kern-.08em
    T\kern-.1667em\lower.7ex\hbox{E}\kern-.125emX}}
\begin{document}

\title{MedXAI: A Retrieval-Augmented and Self-Verifying Framework for Knowledge-Guided Medical Image Analysis}
\author{
\IEEEauthorblockN{Midhat Urooj, Ayan Banerjee, Farhat Shaikh, Kuntal Thakur, Ashwith Poojary, Sandeep Gupta}
\IEEEauthorblockA{\textit{Impact Lab} \\
\textit{Arizona State University}\\
Tempe, AZ, USA \\
Emails: \{murooj, abanerj3, fshaik12, kthakur9, apoojar4, sandeep.gupta\}@asu.edu}
}

\maketitle

\begin{abstract}
Accurate and interpretable image-based diagnosis remains a fundamental challenge in medical AI, particularly under domain shifts and rare-class conditions. Deep learning models often struggle with real-world distribution changes, exhibit bias against infrequent pathologies, and lack the transparency required for deployment in safety-critical clinical environments. We introduce \textbf{MedXAI} (An Explainable Framework for Medical Imaging Classification), a unified expert knowledge based framework that integrates deep vision models with clinician-derived expert knowledge to improve generalization, reduce rare-class bias, and provide human-understandable explanations by localizing the relevant diagnostic features rather than relying on technical post-hoc methods (e.g., Saliency Maps, LIME).

We evaluate MedXAI across heterogeneous modalities on two challenging tasks: (i) Seizure Onset Zone localization from resting-state fMRI, and (ii) Diabetic Retinopathy grading. Experiments on ten multicenter datasets show consistent gains, including a 3\% improvement in cross-domain generalization and a 10\% improvmnet in F1 score of rare class, substantially outperforming strong deep learning baselines. Ablations confirm that the symbolic components act as effective clinical priors and regularizers, improving robustness under distribution shift. MedXAI delivers clinically aligned explanations while achieving superior in-domain and cross-domain performance, particularly for rare diseases in multimodal medical AI.
\end{abstract}

\section{Introduction}
Medical imaging is central to disease diagnosis and treatment planning in conditions such as diabetic retinopathy (DR), tumor detection, and neurodegenerative disorders. While deep learning (DL) models, particularly Convolutional Neural Networks (CNNs) and Vision Transformers (ViTs), have achieved remarkable predictive performance \cite{dosovitskiy2020,simonyan2015}, three key challenges limit their adoption in real-world clinical practice: (i) \textbf{interpretability}, as DL models are often black boxes and post-hoc explainability methods such as Grad-CAM \cite{selvaraju2017} and SHAP \cite{lundberg2017} remain heuristic, static, and disconnected from clinical reasoning. Attention or uncertainty based methods \cite{wang2021,volpi2018} provide partial insight but do not leverage structured medical knowledge, while reinforcement learning and meta-learning approaches \cite{mnih2015} allow adaptive predictions but lack clinically grounded explanations. Existing model explainability in medical AI often uses technical terminology that does not align with clinical language, making it difficult for healthcare professionals and patients to interpret.
(ii) \textbf{rare-class learning}, because clinically significant pathologies are often infrequent and heterogeneous, causing traditional DL models to underperform in capturing nuanced visual and clinical patterns of minority disease classes \cite{Liang2021HumanCenteredAI}; and (iii) \textbf{cross-domain generalization}, as models trained on one institution’s data frequently fail on data from other centers due to variations in acquisition protocols, imaging devices, or patient demographics \cite{zhou2022,wu2022,Gulrajani2020InSearch}.
\begin{figure}[t]
    \centering
    \includegraphics[width=0.8\columnwidth]{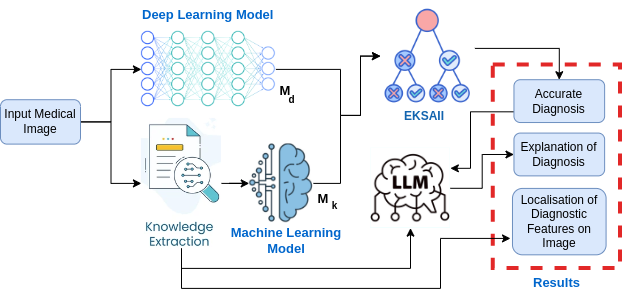} 
    \caption{Conceptual overview of the MedXAI framework. 
    Knowledge extraction is based on a Retrieval-Augmented and Self-Verifying 
    Framework through LLM.}
    \label{fig:neuro_guard_framework}
\end{figure}

Rule-based and expert knowledge systems offer interpretability but struggle to scale across heterogeneous populations and imaging protocols \cite{Boerwinkle2020fMRI, Lee2014fMRI, Calisto2021HumanCentricAI, Cai2021HumanCenteredTools}. expert knowledge based learning, which combines DL feature extraction with symbolic reasoning, has emerged as a promising solution \cite{Han2020UnifyingNeural, Ozkan2020TrainingAcross}. These systems leverage neural networks to capture complex representations while encoding domain knowledge and logical constraints to ensure clinically consistent reasoning. Yet, existing expert knowledge based approaches rarely address rare-class bias, intra-class variability, and cross-domain generalization in a unified framework.

To address these limitations, we propose MedXAI, a expert knowledge based framework that seamlessly integrates structured clinical knowledge with deep neural representations in a scalable and interpretable manner. Clinical expertise is extracted from Pubmed fetcher through an RAG connected with an LLM in the knowledge extractor module. The framework combines: (i) a data-driven neural branch that captures complex imaging features, and (ii) a knowledge-informed symbolic branch that encodes clinically derived rules. An adaptive routing mechanism inspired by Hunt’s algorithm constructs a decision tree of expert models, each specialized for a specific class and drawing from both neural and symbolic branches. The resulting diagnosis is then processed by a large language model (GPT-4), which also receives the symbolic knowledge features. GPT-4 generates a clinically aligned, fact-based explanation in human-understandable language, bridging the gap between technical model outputs and interpretable, actionable insights for medical practitioners and patients as shown in Figure ~\ref{fig:neuro_guard_framework}.

We validate MedXAI on two clinically significant tasks: Seizure Onset Zone (SOZ) localization from MRI and DR grading from retinal fundus images. Experiments on ten multicenter datasets show consistent improvements over state-of-the-art DL baselines, achieving a 10\% improved accuracy in rare-class F1 score. MedXAI not only provides robust predictions under domain shifts but also produces interpretable outputs aligned with clinical reasoning, highlighting relevant anatomical and pathological features.
\section{Proposed Method}

The architecture comprises two complementary branches: (1) a \textbf{Deep Learning (DL) branch}, $f_{\text{DL}}: \mathcal{X} \rightarrow \mathcal{Y}$, which learns hierarchical representations from raw inputs, and (2) an \textbf{Expert Knowledge Processor (EKP)}, $f_{\text{KL}}: F^* \rightarrow \mathcal{Y}$, which encodes structured clinical knowledge. These branches are combined using the \textbf{Expert Knowledge and Supervised AI Integration (EKSAII)} algorithm, which maps a structured, human-interpretable knowledge vector $F^* \in \mathbb{R}^{m}$ to the output space. As shown in Figure~\ref{fig:neuro_guard_framework}, the framework integrates a deep learning model ($M_d$) for feature extraction and a knowledge-based model ($M_k$) derived from expert rules. The outputs are fused via EKSAII to inform a large language model (LLM), which generates the final interpretable results: accurate diagnosis, explanation of the diagnosis, and localization of diagnostic features.

\subsection{Workflow}

Raw input images and disease classification details are first processed by medical experts, who encode clinical knowledge into structured expert knowledge representations. This expert knowledge are either converted into fixed rules (for simple, independent conditions) or, if complex and interdependent, into expert knowledge features $\mathcal{K} = \{r_1, \dots, r_n\}$ for training a classifier solely on knowledge. These features capture domain-specific concepts ~\emph{such as gray matter in SOZ IC, retinal lesions or blood clots, lesion counts, clinical attributes, ECG morphology, etc}. Both rule-based and knowledge-driven classifiers are collectively referred to as $M_k$ and are entirely knowledge-driven.  

Simultaneously, raw images are processed by a customized deep learning model $M_d$. The outputs of $M_d$ and $M_k$ are then combined using the EKSAII algorithm that is presented in recent work of expert-guided medical AI decision systems ~\cite{Kamboj2024}.  to produce a unified representation that result final classification result, which is subsequently used to generate clinically interpretable predictions and explanations via the LLM.

\subsection{Mathematical Grounding of EKSAII Algorithm}

To manage rare classes and high intraclass variability, MedXAI employs EKSAII Algorithm  ~\cite{Kamboj2024} that selects from a classifier pool $\mathcal{M} = \{M_1, \dots, M_k\}$. The selection is guided by two metrics. First, to quantify a classifier's impact on rare class separability, we introduce the \textbf{Entropy Imbalance Gain (EIG)}. This is derived from the local density $\lambda(x_i)$ of an instance $x_i$ within its $K$-nearest intraclass neighbors $Q(x_i)$:
\small\begin{equation}
\lambda(x_i) = \frac{1}{|Q(x_i)|} \sum_{x_j \in Q(x_i)} \text{dist}(x_i, x_j)^{-1}.
\end{equation}
The normalized density $\gamma(x_i)$ and class entropy $\theta_r$ for a class $c_r$ are:
\scriptsize \begin{equation}
\gamma(x_i) = \frac{\lambda(x_i)}{\sum_{x_k \in c_r} \lambda(x_k)}, \quad \text{and} \quad \theta_r = - \sum_{x_i \in c_r} \gamma(x_i) \log_2 \gamma(x_i).
\end{equation}
\normalsize
The EIG for a classifier $M_d$ is the reduction in entropy imbalance $\eta$ relative to the raw data representation $\eta_R$:
\scriptsize \begin{equation}
\text{EIG}(M_d) = \eta_R - \eta_{M_d}, \quad \text{where} \quad \eta_{M_d} = \max_{c_r} \theta_{M_d,r} - \mathbb{E}[\theta_{M_d,r}].
\end{equation}
\normalsize
A higher EIG signals an improved representation of rare classes. Second, we measure heterogeneity within a predicted partition $s$ using the Gini index, $Gini(s) = 1 - \sum p_i^2$, where high impurity ($Gini(s) > \tau_g$) motivates cascading a subsequent classifier to resolve the partition. 
The adaptive selection process (EKSAII Algorithm) recursively partitions data by selecting the classifier with maximum EIG. The framework is trained end-to-end, and during inference, this algorithm is invoked for ambiguous instances. The final prediction $y_{\text{final}}$ is determined by a fully trained decision tree. The knowledge feature $\mathcal{K}$  and final diagnosis $y_{\text{final}}$ are fed into a large language model (we use GPT-4, though any state-of-the-art LLM could be used). The model generates a human-understandable explanation based on the diagnostic results, knowledge attributes, and clinical facts provided as prior rules in the prompt.

\section{Experiments and Results}
We validated the MedXAI framework across two distinct, high-stakes clinical applications: Seizure Onset Zone (SOZ) localization for epilepsy surgery planning and Diabetic Retinopathy (DR) grading for ophthalmology.

\subsection{Application 1: SOZ Localization in Epilepsy}
\paragraph{Implementation.} For SOZ localization from rs-fMRI, the DL branch was a 2D CNN trained to classify Independent Components (ICs) as noise or non-noise. The EKIE branch was engineered to extract four neurophysiologically-grounded features: number of clusters (K-NumC), ventricular activation (K-ThruV), and temporal sparsity (K-SparseA/F). Given the rarity of SOZ ICs (approx. 5 per subject), we employed SMOTE on the 4-D feature space of the EKIE branch to create a balanced training set. The adaptive selection algorithm was instantiated by first calculating the EIG for each branch, yielding $\text{EIG(EKIE)} = 0.22$ and $\text{EIG(DL)} = 0.027$. Consequently, the EKIE branch was chosen as the primary classifier, with its partitions subsequently refined by the DL branch as dictated by the Gini impurity.

\begin{figure*}[t]  
    \centering
    \includegraphics[width=\textwidth]{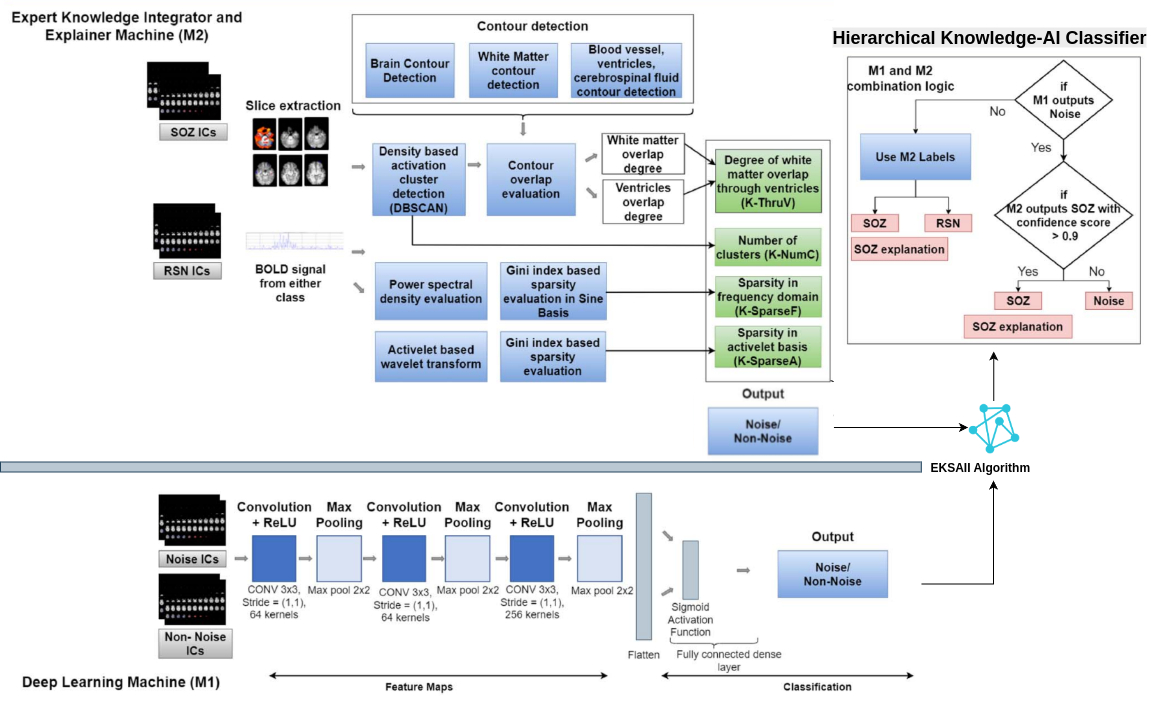} 
    \caption{
        \textbf{DeepXSOZ: A Hybrid Knowledge-AI Architecture for Seizure Onset Zone (SOZ) Localization.} 
        The framework employs a bipartite training architecture to classify Independent Components (ICs) derived from resting-state fMRI (rs-fMRI). The \textbf{Deep Learning Machine} ($\text{M}_{\text{D}}$) as $M_1$ is trained on rs-fMRI ICs for an initial Noise/Non-noise component discrimination. Concurrently, the \textbf{Expert Knowledge Integrator and Explainer Machine} ($\text{M}_{\text{k}}$) as $M_2$ computes a set of expert-derived knowledge components and learns the optimal weight configurations necessary for robust SOZ/RSN (Resting State Network) distinction and the generation of localized classification explanations. During inference, the final SOZ classification is determined by integrating the labels from both $\text{M}_{\text{D}}$ and $\text{M}_{\text{K}}$ via EKSAII algorithm, yielding a final, integrated, and explainable diagnostic result.
    }
    \label{fig:deepxsoz_framework}
\end{figure*}

\paragraph{Results: Rare Class Efficacy and Generalization.} The MedXAI framework proved highly effective in this rare-class detection scenario. As shown in Table~\ref{tab:soz_results}, the integrated approach achieved 84.6\% accuracy and 89.7\% sensitivity, significantly outperforming the standalone DL branch and a knowledge-based baseline (EPIK) done by clinical expert only. This high performance on the rare SOZ class directly enabled a critical clinical outcome: reducing the manual expert review effort from over 110 ICs to just 18 (an 84.2\% reduction). To validate generalization, the model trained on Phoenix Child Health Center (PCH) data was tested on a new, unseen dataset from a different center University
of North Carolina (UNC) without any fine-tuning. The framework's performance remained robust, achieving a statistically equivalent accuracy of 87.5\%. Notably, even as the DL branch's noise-classification accuracy dropped from 80\% to 70\% on the new domain, the EKIE branch compensated for this shift, underscoring how expert knowledge integration is instrumental for mitigating data leakage and ensuring robust generalization. ~\emph{The textual explanation is also generated for each result which is verified by medical experts}.

\begin{table}[h]
\centering

\footnotesize
\begin{tabular}{lccc}
\toprule
\textbf{Method} & \textbf{Acc (\%)} & \textbf{Sens (\%)} & \textbf{Effort} \\
\midrule
DL Branch (2D CNN) & 46.1 & 48.9 & 10 \\
EPIK (Knowledge Baseline) & 75.0 & 79.5 & 43 \\
\textbf{MedXAI (Ours)} & \textbf{84.6} & \textbf{89.7} & \textbf{18} \\
\bottomrule
\end{tabular} \newline
\caption{SOZ localization performance. The fused MedXAI model significantly outperforms individual components and baselines.}

\label{tab:soz_results}
\end{table}

\subsection{Application 2: Diabetic Retinopathy Grading}
\begin{figure*}[t]  
\centering
    \includegraphics[width=\textwidth]{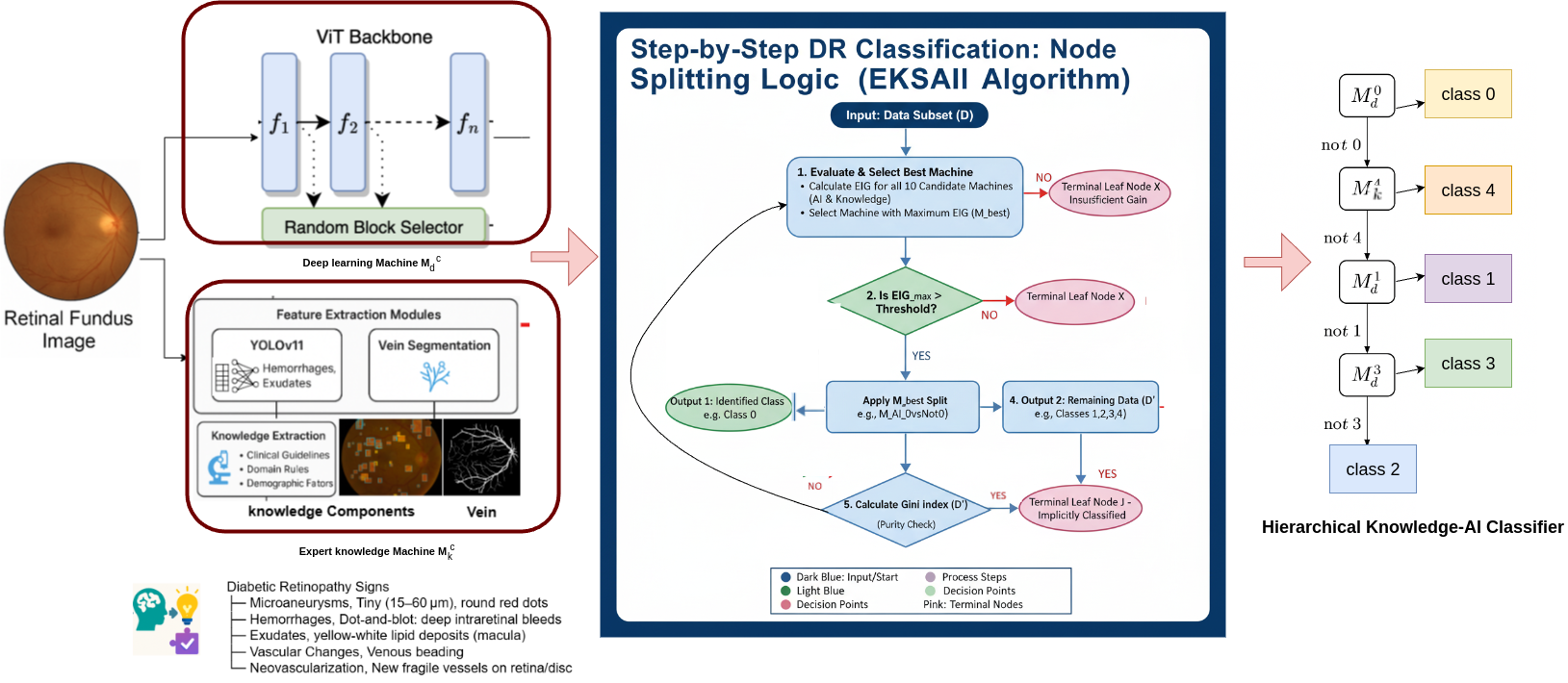} 
    \caption{ 
        The system integrates a \textbf{Deep Learning Machine} ($M_d^c$, ViT backbone for each class c) and an \textbf{Expert Knowledge Machine} ($M_k^c$, clinical features/guidelines for each class c) within a decision tree. 
        The \textbf{EKSAII Algorithm} iteratively selects the optimal binary classifier (maximum Entropy Imbalance Gain, EIG) for node splitting, achieving $84\%$ accuracy across 5 DR classes through an orchestrated, sequential classification path.
    }
    \label{fig:dr_framework}
\end{figure*}

\paragraph{Implementation.} For the 5-class DR grading task, we instantiated the classifier pool $\mathcal{M}$ with ten binary (one-vs-rest) classifiers: five deep learning (ViT-based) branches and five knowledge based branches, each machine specialized for a single DR grade. These ten models were organized into a decision tree following EKSAII algorithm, which generated the final classification. This approach achieved a peak accuracy of 84\% (see Figure ~\ref{fig:dr_framework} for the decision tree structure). Performance was evaluated against strong baselines across four public datasets: APTOS, EyePACS, and Messidor-1/2.

\noindent
The proposed final model tree, comprising both $M_k$ and $M_d$, introduces a novel architectural paradigm for robust 5-class Diabetic Retinopathy (DR) classification, overcoming the limitations of purely data-driven or purely knowledge-based approaches. The system achieves an accuracy of 84\% and is structured as a \textbf{tripartite orchestration} of ten specialized binary classifiers, deployed via a decision tree. As illustrated in Figure~\ref{fig:dr_framework}, the core components consist of two machine types: 

\textbf{(i) Deep Learning Machines ($M_d$)}, powered by a \textbf{Vision Transformer (ViT) backbone} (e.g., DeiT, CvT), which extract high-dimensional abstract features ($f_1, \dots, f_n$); and  
\textbf{(ii) Expert Knowledge Machines ($M_k$)}, implemented as XGBoost classifiers (selected based on Ablation Study 1), which leverage \textbf{interpretable clinical features} (e.g., hemorrhages and exudates segmented by YOLOv11) along with formalized \textbf{clinical guidelines} and demographic factors.  

A key innovation lies in the \textbf{EKSAII (Expert Knowledge-Sensitive AI Integration) Algorithm}, which governs the construction of the decision tree. At each node, EKSAII quantitatively evaluates the splitting efficacy of all candidate $M_d$ and $M_k$ classifiers using the \textbf{Entropy Imbalance Gain (EIG)} metric. This ensures that the most informative and contextually appropriate machine whether AI-derived or expert-defined is selected for the current data subset.  

The resulting tree reflects a \textbf{knowledge-informed classification sequence}: an initial triage by $M_d^0$ (ViT) for ``0 vs. Not 0'' is followed by a strategic alternation between ViT-based classifiers for general pattern recognition ($M_d^1, M_d^3$) and knowledge-based classifiers for clinically salient distinctions ($M_k^4, M_k^2$). This hierarchical, gain-optimized integration of complementary AI and expert knowledge modules provides a diagnostically sound, transparent, and interpretable framework for clinical decision support in DR grading.

\paragraph{Results: Rare Class Detection and Generalization.} The primary benefit of our hierarchical approach was a significant improvement in detecting rare classes. As detailed in Table ~\ref{tab:rare_class_f1}, the full MedXAI framework boosted the F1-score for severe DR grades (3 and 4) by over 10\% compared to a standalone DL model.

\begin{table}[h]  
\centering

\footnotesize
\begin{tabular}{lcc}
\toprule
\textbf{Method} & \textbf{F1 (Grade 3)} & \textbf{F1 (Grade 4)} \\
\midrule
DL Branch (ViT) & 45.2 & 51.8 \\
MedXAI (Fusion) & \textbf{50.1 (+10.8\%)} & \textbf{57.3 (+10.6\%)} \\
\bottomrule
\end{tabular} \newline
\caption{F1-Score (\%) for rare DR classes (MDG setting). MedXAI significantly boosts detection performance.}

\label{tab:rare_class_f1}
\end{table}

\paragraph{Single-Domain Generalization (SDG).} As summarized in Table ~\ref{tab:sdg_summary}, the MedXAI framework consistently outperformed specialized ViT-based baselines in three of the four SDG settings. For instance, when trained on APTOS and test on (EyePACS, MESSIDOR1 and MESSIDOR2), our unweighted fusion strategy achieved an average cross-domain accuracy of 59.9\%, surpassing the best baseline (58.6\%). Similarly, when trained on MESSIDOR2 and test on (EyePACS, MESSIDOR1 and Aptos), the weighted fusion achieved 65.5\% accuracy, underscoring the framework's robustness. This confirms that symbolic knowledge provides a strong inductive bias that aids generalization from limited source data.

\begin{table}[t]  
\centering
\resizebox{\columnwidth}{!}{%
\begin{tabular}{lcccc}
\toprule
\textbf{Source} & \textbf{DL (ViT)} & \textbf{EKIE} & \textbf{MedXAI (Fusion)} & \textbf{Best Baseline} \\
\midrule
APTOS & 53.9 & 56.6 & \textbf{59.9} & 58.6 (SD-ViT) \\
MESSIDOR & 57.0 & \textbf{67.1} & \textbf{67.1} & 55.9 (SPSD-ViT) \\
MESSIDOR2 & 41.1 & 65.2 & \textbf{65.5} & 62.1 (SPSD-ViT) \\
EYEPACS & 50.6 & 60.1 & 61.7 & \textbf{62.5} (SPSD-ViT) \\
\bottomrule
\end{tabular}
}\newline
\caption{Single-Domain Generalization (SDG) performance comparison (Accuracy \%), where a model is
trained on one domain and tested on the others,}

\label{tab:sdg_summary}
\end{table}

\paragraph{Multi-Domain Generalization (MDG).} In the more comprehensive MDG setting (Table ~\ref{tab:mdg_results}), our MedXAI achieves average 67.95\% accuracy, outperforming numerous complex DG methods and the standalone ViT-based DL branch (61.2\%). The strong performance of our knowledge-centric components validates their critical role in achieving robust generalization across diverse clinical environments.

\begin{table}[h]
\centering

\resizebox{\columnwidth}{!}{%
\begin{tabular}{llccccc}
\toprule
\textbf{Method} & \textbf{Backbone} & \textbf{Aptos} & \textbf{Eyepacs} & \textbf{Messidor} & \textbf{Messidor 2} & \textbf{Avg.} \\
\midrule
Fishr & ResNet50 & 47.0 & 71.9 & 63.3 & 66.4 & 62.2 \\
SPSD-ViT & T2T-14 & 50.0 & 73.6 & 65.2 & 73.3 & 65.5 \\
\midrule
DL Branch (ViT) & DeiT-Small & 50.1 & 69.4 & 58.1 & 67.1 & 61.2 \\
EKIE Branch & Knowledge & \textbf{60.7} & 68.5 & 58.7 & 67.7 & 63.7 \\
MedXAI (Fusion) & ViT+EKIE & 53.1 & \textbf{74.8} & \textbf{68.3} & \textbf{75.6} & \textbf{67.95} \\
\bottomrule
\end{tabular}%
} \newline
\caption{Multi Domain Generalization (MDG) performance comparison (Accuracy \%) where the model trains on three domains and is tested on the held-out one.}

\label{tab:mdg_results}
\end{table}

\subsection{Ablation Studies}

To analyze the contributions of neural and symbolic components and evaluate the reliability of lesion-based biomarkers, we conducted three complementary ablation studies using APTOS as the source domain.

\textbf{Study I: Neural vs. Symbolic vs. expert knowledge based Fusion:} We first assessed the generalization of different model configurations, train on Aptos and test on unseen target domains EyePACS, Messidor-1, and Messidor-2. Table~\ref{tab:aptos_ablation} summarizes results. As you can see Vision Transformer (ViT) alone achieves moderate generalization (average 66.6\%). Symbolic reasoning using lesion-level features (KL) improves average accuracy to 66.4\%, demonstrating the value of structured clinical priors. expert knowledge based integration further improves performance, with non-weighted fusion achieving the highest average accuracy of 72.8\%, confirming that combining neural and symbolic reasoning enhances robustness under domain shift.

\begin{table}[h]
\centering
\resizebox{\columnwidth}{!}{
\begin{tabular}{lccc}
\toprule
\textbf{Setting} & \textbf{EyePACS} & \textbf{Messidor-1} & \textbf{Messidor-2} \\
\midrule
Neural Only (ViT) & 66.6 & 46.4 & 48.9 \\
Symbolic Only (KL) & 66.4 & 49.6 & 53.9 \\
Neural + Symbolic (Non-Weighted) & \textbf{72.8} & \textbf{50.6} & \textbf{54.3} \\
Neural + Symbolic (Weighted) & 67.4 & 49.6 & 53.9 \\
\bottomrule
\end{tabular}
}
\newline\caption{Performance of neural, symbolic, and fused models trained on APTOS and tested on unseen domains. expert knowledge based fusion achieves the best generalization.}
\label{tab:aptos_ablation}
\end{table}
\textbf{Study II: Evaluating Lesion Biomarkers Across Classifiers.}
Next, we examined the discriminative power of four lesion biomarkers 
\textit{exudates, hard hemorrhages, soft hemorrhages, and cotton wool spots} 
using multiple classifiers. Table~\ref{tab:ml_models} reports performance on 
APTOS. Gradient Boosting achieves the highest accuracy (84.65\%). confirming the strong predictive value of lesion-only symbolic 
features. Based on these results, Gradient Boosting is selected as the primary 
symbolic classifier for subsequent experiments.
 
\begin{table}[H]
\centering
\renewcommand{\arraystretch}{0.85}
\resizebox{0.8\columnwidth}{!}{
\begin{tabular}{lcc}
\toprule
\textbf{Model} & \textbf{Accuracy} & \textbf{F1-Score} \\
\midrule
Logistic Regression & 0.773 & 0.732 \\
SVM & 0.781 & 0.743 \\
Random Forest & 0.817 & 0.812 \\
Gradient Boosting & \textbf{0.847} & \textbf{0.841} \\
K-Nearest Neighbors & 0.781 & 0.790 \\
\bottomrule
\end{tabular}
}
\newline
\caption{Performance comparison of classifiers using lesion biomarkers on APTOS.}
\label{tab:ml_models}
\end{table}

\textbf{Study III: Impact of Retinal Vein Features.}
Finally, we tested whether adding retinal vein morphology features 
(tortuosity, caliber, branching angles) improves classification. 
Table~\ref{tab:vein_ablation} shows results using Gradient Boosting 
(best model from Study II). Incorporating vein features reduces performance, 
indicating that vascular measurements introduce domain-sensitive variability. 
Lesion-only biomarkers remain the most robust and interpretable symbolic inputs.

\begin{table}[H]
\centering
\renewcommand{\arraystretch}{0.85}
\resizebox{0.75\columnwidth}{!}{
\begin{tabular}{lcc}
\toprule
\textbf{Feature Set} & \textbf{Accuracy} & \textbf{F1-Score} \\
\midrule
Lesions Only & \textbf{0.847} & \textbf{0.841} \\
Lesions + Vein & 0.725 & 0.739 \\
\bottomrule
\end{tabular}
}
\newline
\caption{Gradient Boosting performance with and without vein-based features.}
\label{tab:vein_ablation}
\end{table}

\section{Conclusion}
This work introduces MedXAI, a unified neuro-symbolic framework designed to overcome the critical barriers of interpretability, rare-class bias, and cross-domain generalization that currently limit the deployment of deep learning in safety-critical clinical environments. The integration of Large Language Models (LLMs) further ensures that diagnostic outputs are accompanied by human-understandable, fact-based explanations, bridging the communication gap between AI systems and medical practitioners.

Extensive validation on ten multicenter datasets across two distinct tasks—Seizure Onset Zone localization and Diabetic Retinopathy grading, demonstrates the framework's robustness. MedXAI achieved a 10\% improvement in F1 scores for rare disease classes and reduced manual expert review effort by over 84\%, validating its utility in high-stakes, real-world settings. Furthermore, the system exhibited superior generalization capabilities when tested on unseen domains, confirming that symbolic knowledge acts as a powerful prior to mitigate data leakage and distribution shifts.

\subsection{Future Work}
Future research will focus on extending the MedXAI framework on heterogeneous medical data of different modalities, such as integrating high-dimensional genomic profiles with continuous time-series data (e.g., EEG or real-time vitals).

\bibliographystyle{IEEEtran}
\bibliography{ref}  
\end{document}